\documentclass[letterpaper, 10 pt, conference]{ieeeconf}  

\IEEEoverridecommandlockouts                              

\usepackage{graphicx} 
\usepackage{amsmath} 
\usepackage{amssymb}  

\title{\LARGE \bf
Exploiting Physical Human-Robot Interaction to Provide a Unique Rolling Experience with a Riding Ballbot
}

\author{Chenzhang Xiao$^{1}$*, Seung Yun Song$^{1}$, Yu Chen$^{1}$, Mahshid Mansouri$^{1}$, Jo\~{a}o  Ramos$^{1}$, Adam W. Bleakney$^{2}$, \\ William R. Norris$^{3}$, and Elizabeth T. Hsiao-Wecksler$^{1}$
\thanks{This work was supported by NSF Grant No. 2024905.}
\thanks{$^{1}$Mechanical Science and Engineering, University of Illinois at Urbana-Champaign (UIUC), Urbana, IL 61801 USA (cxiao3, ssong47, yuc6, mm64, jlramos, ethw)@illinois.edu. *Corresponding author.
        }%
\thanks{$^{2}$Disability Resources and Educational Services, UIUC, Champaign, IL 61820 USA (bleakney@illinois.edu)}%
\thanks{$^{3}$Industrial and Enterprise Systems Engineering, UIUC, Champaign, IL 61820 USA (wrnorris@illinois.edu)}}%

\begin{document}

\maketitle
\thispagestyle{empty}
\pagestyle{empty}

\begin{abstract}

This study introduces the development of hands-free control schemes for a riding ballbot, designed to allow riders including manual wheelchair users to control its movement through torso leaning and twisting. The hardware platform, Personal Unique Rolling Experience (PURE), utilizes a ballbot drivetrain, a dynamically stable mobile robot that uses a ball as its wheel to provide omnidirectional maneuverability. To accommodate users with varying torso motion functions, the hanads-free control scheme should be adjustable based on the rider's torso function and personal preferences. Therefore, concepts of (a) impedance control and (b) admittance control were integrated into the control scheme. A duo-agent optimization framework was utilized to assess the efficiency of this rider-ballbot system for a safety-critical task: braking from 1.4 m/s. The candidate control schemes were further implemented in the physical robot hardware and validated with two experienced users, demonstrating the efficiency and robustness of the hands-free admittance control scheme (HACS). This interface, which utilized physical human-robot interaction (pHRI) as the input, resulted in lower braking effort and shorter braking distance and time. Subsequently, 12 novice participants (six able-bodied users and six manual wheelchair users) with different levels of torso motion capability were then recruited to benchmark the braking performance with HACS. The indoor navigation capability of PURE was further demonstrated with these participants in courses simulating narrow hallways, tight turns, and navigation through static and dynamic obstacles.  By exploiting pHRI, the proposed admittance-style control scheme provided effective control of the ballbot via torso motions. This interface enables PURE to provide a personal unique rolling experience to manual wheelchair users for safe and agile indoor navigation.

\end{abstract}

\section{INTRODUCTION}

Wheeled assistive devices provide rolling mobility to people with lower extremity disabilities; however, they often fail to address the functional and mobility needs of these users. The manual wheelchair requires the use of both hands for propulsion and often results in upper extremity injuries due to repetitive motions \cite{nichols1979wheelchair,bayley1987wheelchair,gellman1988wheelchair}. While powered wheelchairs with motorized wheels controlled by a joystick eliminate the physical upper-body exertion, they are heavier, more expensive, require modifications to personal vehicles for transportation, and may not be suitable for navigation in constrained indoor environments due to their bulky size and lack of agility. Two-wheeled self-balancing wheelchairs, inspired by Segway, offer improved agility and allow users to control forward and backward movements through torso leaning \cite{omeo}. However, their size and lack of omnidirectional maneuverability still pose challenges in indoor spaces.

\begin{figure}[thpb]
  \centering
  \includegraphics[scale=0.5]{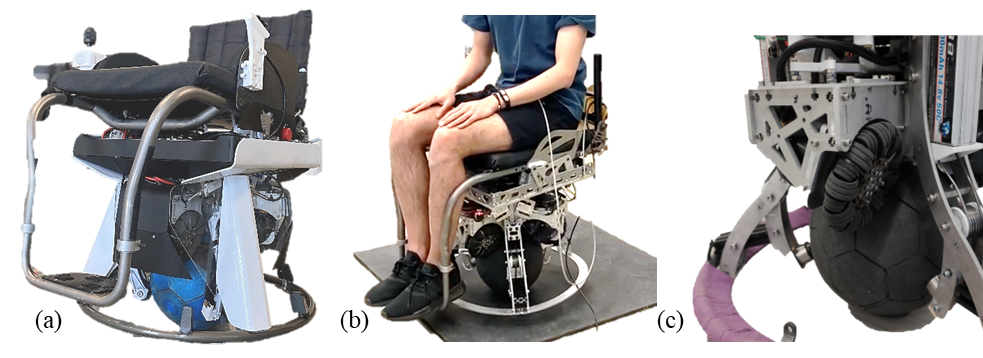}
  \caption{(a) The riding ballbot, PURE, with its shroud. (b) A rider sitting on PURE and balancing. (c) A close-up image of omniwheel and ball utilized in the ballbot drivetrain.}
  \label{overview}
\end{figure}

We have developed a ballbot drivetrain that has omnidirectional maneuverability, high agility, minimal footprint, and high payload capacity while maintaining dynamic stability \cite{xiao2023design}. It is powered by three omniwheel-motor pairs. In this paper, we present a novel mobility device based on this drivetrain, named PURE (Personal Unique Rolling Experience) (Fig. \ref{overview}). The minimal footprint and high agility allow for easy maneuverability in constrained indoor environments. Due to its self-balancing design, PURE can achieve omnidirectional motion by leaning the torso in any direction; thus allowing seated mobility that frees the hands to do other simultaneous activities such as carrying objects in both hands.
However, the torso lean control of the ballbot drivetrain shares the same issues as the two-wheeled self-balancing wheelchairs: ineffective shifting of center-of-mass (COM) from torso leaning compared with the whole-body leaning, and the lack of tuning capability based on the torso function capability of individual riders. New hands-free control schemes must be developed to allow for effective torso lean control of the ballbot drivetrain, with tunable parameters that personalize the system response to each rider’s torso function and range of motion.

The torso lean control of a ballbot can be generalized to a human-robot collaborative task with physical human-robot interactions \cite{alami2006safe, de2008atlas}. In this research field, impedance and admittance control schemes are commonly used in tasks that require such physical interactions and collaboration \cite{hogan1984impedance,ikeura1995variable}. An impedance controller regulates forces based on deviations from a set point, allowing the system to adjust its resistance to externally applied interaction forces and moments. This approach has been widely applied in areas such as haptics control \cite{agravante2014collaborative, yang2017interface}, rehabilitation robots \cite{yu2015human}, teleoperation \cite{love2004force, duchaine2007general}, and co-manipulation tasks\cite{ficuciello2014cartesian, ficuciello2015variable}. Conversely, an admittance controller modulates motion in response to measured forces and moments, rendering the desired system impedance or admittance in physical interactions. Admittance control has been utilized in applications such as control of the exoskeletons \cite{gui2017toward,li2018physical}, end-effector interaction for power-amplification or load reduction \cite{kazerooni1993human, colgate2003intelligent}, and interactions with mobile robots \cite{li2012passivity, augugliaro2013admittance, wang2015stability}. Both control schemes can be adapted for the torso lean control of a ballbot, and are compatible with the cascaded LQR-PI controller implemented in our ballbot drivetrain. However, there has been no prior research exploring these two control schemes in the context of a dynamically stable riding device, nor any evidence of coupled stability \cite{keemink2018admittance} between the rider and ballbot using these two control schemes. 

To address these challenges, we proposed to investigate these two groups of hands-free control schemes and to evaluate their performance in a safety-critical task of PURE riding: braking from 1.4 m/s to full stop. We conducted studies using both simulation and physical robot experiments, involving experienced and novice able-bodied and manual wheelchair users. Various metrics, including braking effort, braking time, braking distance, etc., were utilized to assess the braking performance. We hypothesized the control scheme based on admittance control, which exploits the physical human-robot interactions to render desired system behavior, would be superior compared with pure impedance-based control schemes.

The rest of this paper is structured in the following way. Section II introduces the rider-ballbot model, incorporating torso motion in the system dynamics. Section III  presents two groups of hands-free control schemes and Section IV details a dual-agent simulation study to obtain insights into controller performances during the braking task. Section V describes two physical robot experiments: (1) evaluate the braking effectiveness of all proposed controllers with two experienced riders, and (2) benchmark braking performance and validate the feasibility of indoor navigation with 12 novice riders (six able-bodied and six manual wheelchair users). Sections VI and VII present the Discussion and Conclusions of the study.

\section{MODELING OF RIDER-BALLBOT SYSTEM}

\begin{figure}[thpb]
  \centering
  \includegraphics[scale=0.45]{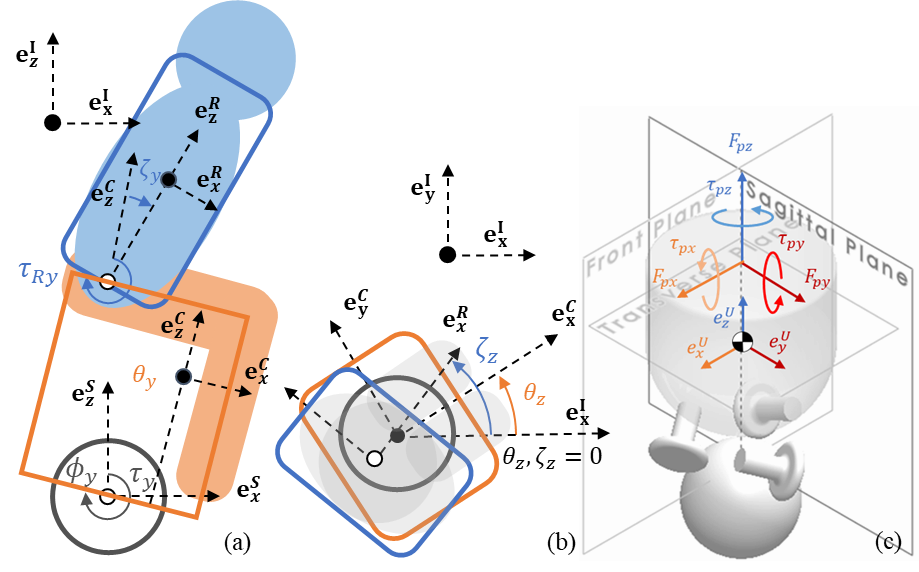}
  \caption{Modeling of rider-ballbot system in the (a) sagittal plane, (b) transverse plane, and (c) model that isolates the ballbot from the rider-ballbot model with physical human-robot interactions.}
  \label{model}
\end{figure}

Rider-ballbot models were developed to investigate the coupled dynamics of the rider and the ballbot drivetrain, especially the effect of torso leaning. These models can be utilized to analyze and simulate the behavior of the rider-ballbot system during the torso-leaning control. The torso motions of the rider were decomposed into torso leaning in the sagittal and frontal planes, as well as torso twisting in the transverse plane (Fig.\ref{model}). Coordinate frames were assigned to the rider's upper body ($R$), lower body plus chassis ($C$), and the spherical wheel ($S$), in addition to the inertia frame ($I$). In this setup, we can effectively describe the dynamics of torso leaning, while assuming the rider’s lower body is not moving relative to the drivetrain chassis. More details of the model derivation can be found in \cite{xiao2022personal}. Euler-Lagrange's method was used to derive the equations of motion for each planar model \cite{spong2004robot}. The Newtonian method was further used to derive the resultant physical human-robot interactions (pHRI) applied to the center of the ballbot due to rider dynamics and torso leaning \cite{spong2004robot}.

\section{CONTROL SYSTEM DEVELOPMENT}

\subsection{Translation Control}

This section introduces several variants of hands-free impedance and admittance control schemes to control the translational motion of PURE riding. A baseline LQR-PI controller was previously developed as the low-level control \cite{xiao2023design}. However, using this baseline LQR-PI controller alone for hands-free control lacked adjustability, i.e., the system performance could not be tailored to the rider’s specific abilities or preferences. Additionally, it would tend to produce a resistive behavior, where the chassis tilted in the opposite direction of any translational motions to resist the applied interactions (since all command states are zero). In contrast, a compliant behavior is more desired, when the device tilts towards the direction of the resultant translational direction caused by physical interactions.

\begin{figure}[thpb]
  \centering
  \includegraphics[scale=0.6]{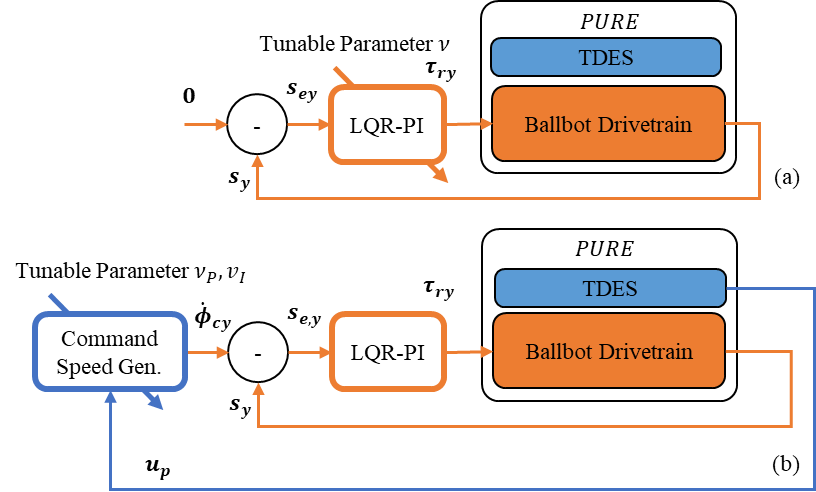}
  \caption{Control block diagram for (a) hands-free impedance control scheme (HICS) and (b) hands-free admittance control scheme (HACS). Both schemes maintain the low-level LQR-PI balancing controller structure. HICS adjusts control gain associated with speed tracking, while HACS provides command speed utilizing measured physical human-robot interaction.}
  \label{ctrl_sys}
\end{figure}

\subsubsection{Impedance Control Scheme}

The hands-free impedance control scheme (HICS) was proposed to address the adjustability and resistive behavior of the baseline controller. A generic impedance controller modifies torque input based on the deviation from the steady state to render the impedance. In the particular case of PURE, we introduced a scaled speed control gain $\hat k_{3j}$ for the translation speed in the baseline LQR-PI controller to adjust system response towards a non-zero speed (Fig. \ref{ctrl_sys}a). The input torque to the reference model in the cascaded LQR-PI controller ($\tau_{rj}$) was then derived to be:
\begin{equation}
\tau_{rj}  =  k_{1j}(0-\theta_j) +  k_{2j}(0-\dot\theta_j) + \hat k_{3j}(0-\dot\phi_j)
\end{equation}
for $j\in[x,y]$ assuming torque in the frontal or sagittal plane, respectively. Two candidate HICS were further proposed: 
\begin{equation}
\begin{aligned}
    \text{HICS1: } & \hat k_{3j} = \nu_j k_{3j} \\
    \text{HICS2: } & \hat k_{3j} = \nu_j \frac{\dot\phi_j}{\dot\phi_{max}} k_{3j}\\
\end{aligned}
\end{equation}
where $\nu_j\in[0,1]$ is a tunable sensitivity parameter between 0 and 1 that can be utilized to adjust system behavior. $\dot\phi_{max}$ is the maximum angular speed of the ball, which was set to 17.6 rad/s (equivalent to a translational speed of 2 m/s). Smaller $\nu_j$ would result in lower speed control gains, and thus less impedance against translational motions.

\subsubsection{Admittance Control Scheme}

In the hands-free admittance control scheme (HACS), we kept all control gains at their optimal values while providing a command speed $\dot\phi_{cj}$ generated from the physical human-robot interactions (pHRI) $\mathbf{u_{p}} = [F_{px},F_{py},F_{pz}, \tau_{px},\tau_{py},\tau_{pz}]^T$ applied to the center of the seat of PURE (Fig. \ref{model}c) as a result of torso leaning.
The reference input torque was defined as:
\begin{equation}
        \tau_{rj} = k_{1j}(0-\theta_j) + k_{2j}(0-\dot\theta_j) +  k_{3j}(\dot\phi_{cj}(\tau_{pj})-\dot\phi_j) \\
\end{equation}
The pHRI moment in sagittal and frontal planes ($\tau_{pj}$) was utilized to generate the rider command speed as it has a better correlation with torso leaning (Fig.\ref{ctrl_sys}b). Three candidate HACSs were proposed:
\begin{equation}
\begin{aligned}
    \text{HACS1:} & \dot\phi_{cj} = \nu_{Pj}\tau_{pj} \\
    \text{HACS2:} & \dot\phi_{cj} = \nu_{Ij} \int{\tau_{pj}}\,dt \\
    \text{HACS3:} & \dot\phi_{cj} = \nu_{Pj}\tau_{pj} + \nu_{Ij} \int{\tau_{pj}}\,dt \\
\end{aligned}
\end{equation}
where $\nu_{Pj}$ and $\nu_{Ij}$ are sensitivity parameters between 0 and 1 that can be tuned to adjust system behavior.

\subsection{Spin Control}

We further utilized the measured torso twist angle $\zeta_z$ as the reference signal for the hands-free control of yaw motion:
\begin{equation}
        \tau_{rz} = \nu_z \bigl( k_{1z}(\zeta_z-\theta_z) +  k_{2z}(0-\dot\theta_z) \bigr) \\
    \label{study2:yaw_control}
\end{equation}
where $k_{1z}$ and $k_{2z}$ are LQR control gains for yaw model, $\nu_z$ is a sensitivity parameter between 0 and 1 adjusting system response for torso twisting.

\section{SIMULATED ROBOT EXPERIMENT}

A simulation study with the rider-ballbot model was conducted to evaluate the effectiveness of controlling with HICS or HACS during a braking task from 1.4 m/s (close to human walking speed \cite{mcneill2002energetics}) to a complete stop. The simulation focused exclusively on sagittal plane motion ($j=y$), as agile navigation primarily occurs in the forward direction.

\subsection{Duo-Agent Simulation Setup}

Duo-agent control policies (ballbot and rider) were implemented separately to simulate the hands-free control of the translational motions of PURE via torso leaning. 
Candidate HICS and HACS algorithms were implemented explicitly for PURE control while sweeping through various sensitivity parameters. The rider agent control policy was obtained inexplicitly by leveraging trajectory optimization tools, due to difficulties in deriving a control law of a human rider. We assumed the rider agent would aim to perform the task with minimal braking effort, which was characterized using the following objective function:

\begin{equation}
    J = \int_{t_0}^{t_f} {\mathbf{\hat{s}_{y}}(t)^T \mathbf{Q}_y \mathbf{\hat{s}_{y}}(t) + \mathbf{\hat{u}_{y}}(t)^T \mathbf{R}_y \mathbf{\hat{u}_{y}}(t)} \ dt
    \label{eqn_brake}
\end{equation}
where $J$ is the braking effort, $t_0$ and $t_f$ are the start and end time of the braking process, $\mathbf{\hat{s}_{y}} = [\zeta_y,\theta_y,\phi_y, \dot{\zeta}_y, \dot{\theta}_y, \dot{\phi}_y]$ is the state vector of the rider-ballbot model, $\mathbf{\hat{u}_{y}} = [\tau_{Ry}, \tau_y]^T$ is the input torque vector for rider torso torque and ballbot drivetrain torque, $\mathbf{Q}_y$ and $\mathbf{R}_y$ are diagonal weight matrices that can be designed on our perception of an efficient braking performance: 1) small torso lean angle, 2) low torso torque, and 3) low braking distance. In this case, the following weight matrices were utilized:
\begin{equation}
    \begin{aligned}
        \mathbf{Q}_y & = diag ( \left(\frac{1}{\zeta_{y_{ROM}}}\right)^2, 0, \left(\frac{1}{\phi_{y_{max}}}\right)^2, \left(\frac{1}{\dot\zeta_{y_{max}}}\right)^2, 0, 0 ) \\
        \mathbf{R}_y & = diag(\left(\frac{1}{\tau_{Ry_{max}}}\right)^2,0)
    \end{aligned}
    \label{eqn_weight}
\end{equation}
where $diag(\cdot)$ denotes the function for generating a diagonal matrix with its principle diagonal elements, $\zeta_{y_{ROM}}, \dot\zeta_{y_{max}}$ are the torso range of motion (ROM) and maximum angular rate, $\phi_{y_{max}}$ is the desired braking distance, and $\tau_{Ry_{max}}$ is the maximum torso torque.

The following trajectory optimization problem was formulated to find optimal state $\mathbf{\hat{s}_{y}^*}(t)$ and input torque $\mathbf{\hat{u}_{y}^*}(t)$ trajectories for each hands-free control scheme with a specific sensitivity parameter  ($\nu_y$ or $\nu_{Py}$).

\begin{equation}
    \begin{aligned}
\min_{\mathbf{\hat{s}_{y}(t),\hat{u}_y(t)}} \quad & J(\mathbf{\hat{s}_y, \hat{u}_y}) & \\
\text{subject to} \quad & \tau_{ry} = f_{ctrl}(\mathbf{s_y}, \nu_y, \nu_{Py}) &  \\
 & \mathbf{\ddot{q}_{y}}(t) = f_{y}(\mathbf{q_{y}}(t), \mathbf{\dot{q}_{y}}(t), \mathbf{\hat{u}_{y}}(t)) & \\
& H(t,\mathbf{\hat{s}_j(t),\hat{u}_y(t)}) \le 0  & \\
& G(t_0,t_F, \mathbf{\hat{s}_y}(t_0),\mathbf{\hat{s}_y}(t_F)) \le 0 & \\
\end{aligned}
\label{eqn_optimization}
\end{equation}
where $f_{ctrl}$ is the control function for HACS or HICS algorithms and $f_y$ is the equation of motion of the rider-ballbot model in the sagittal plane. The path constraints ($H(\cdot)\le0$) and boundary constraints ($G(\cdot)\le0$) were utilized to ensure optimal trajectories to have a steady-state initial state, a fully stopped final state, and remain within state and input boundaries. More details of the formulation of this optimization problem can be found in \cite{xiao2022personal}.

The direct collocation algorithm \cite{kelly2017introduction} was used to obtain the state and input trajectories that minimized the braking effort ($J$) for each hands-free control scheme while varying the respective sensitivity parameter (from 0 to 1 with intervals of 0.1). Due to the nature of the collocation method, HACS algorithms, which involved the integration of the system states, cannot be solved directly; thus HACS-2 and HACS-3 were excluded from this simulation study. For the remaining control schemes, 33 conditions were solved (3 control schemes $\times$ 11 sensitivity parameters per control scheme). The feasibility was determined by whether the algorithm could find a viable solution, and the effectiveness was quantified by the resultant value of the braking effort in (\ref{eqn_optimization}). A smaller braking effort value indicates a more effective control scheme for a given sensitivity parameter.

\begin{figure}[thpb]
  \centering
  \includegraphics[scale=0.5]{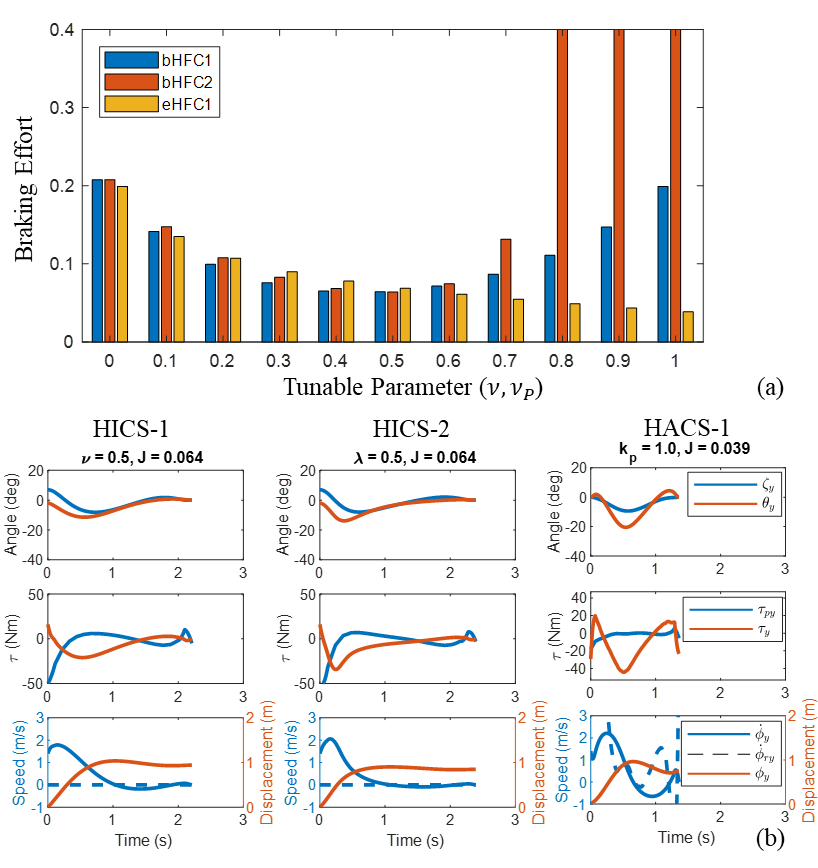}
  \caption{(a) Simulated braking effort ($J$) for each type of control scheme subject to sensitivity parameters ($\nu$ or $\nu_{Py}$) from 0 – 1. (b) State and torque trajectories in each control scheme that produced the lowest braking effort. Top: trajectory of chassis tilt (orange) and torso lean angles (blue). Middle: drivetrain (orange) and torso torques (blue). Bottom: PURE displacement (orange), translational speed (blue), and command speed (dash blue).}
  \label{sim_result}
\end{figure}

\begin{table}[h]
\caption{Results for Simulation Study }
\label{sim_table}
\begin{center}
\begin{tabular}{c c c c c}
Metrics & Symbol & HICS-1 & HICS-2 & HACS-1\\
\hline
Braking Effort (J)  & $J$ & 0.064 & 0.064 & 0.039*\\
Torso ROM  ($^\circ$) & $\zeta_{y,ROM}$    & 14.9 & 14.9 & 9.3*\\
Max pHRI Torque (Nm)& $\tau_{py,max}$      & 50 & 50 & 18*\\
Braking Distance  (m)  & $L$                 & 1.0 & 0.9 & 0.78*\\
Braking Time (sec)    & $T$                 & 2.2 & 2.4 & 1.4*\\
\end{tabular}
\end{center}
\end{table}

\subsection{Results and Discussion}
The direct collocation algorithm successfully obtained optimal state and input trajectories for most simulation cases, except for HICS-2 with higher values of sensitivity parameters (Fig. \ref{sim_result}); demonstrating the theoretical feasibility of the proposed hands-free control schemes. The sensitivity parameter effectively changed the resultant braking effort. For both HICS, the braking effort decreased and then increased as the sensivity parameter $\nu$ increased from 0 to 1. For HACS-1, the braking effort decreased singularly as the $\nu_p$ increased (Fig. \ref{sim_result}a).
Among these control schemes, HACS-1 required the smallest torso lean angle and torso torque, and the shortest braking time and distance to complete the task (TABLE \ref{sim_table}). These results suggested that HACS could be promising to provide an effective torso control of translational motions when riding PURE.

\section{HUMAN SUBJECT TEST WITH PHYSICAL ROBOT}

Human subject tests with physical robot hardware were conducted to evaluate the effectiveness of these control schemes of PURE for the braking task and validate the feasibility of hands-free control of PURE for indoor navigation.

\subsection{Hardware Platform}
To measure applied pHRI, a custom Torso-dynamics Estimation System (TES) was developed \cite{song2024driving}.
The TES was constructed from a manual wheelchair frame instrumented with six load cells to calculate sagittal and frontal plane rider-seat interaction torques, $\tau_{px}$ and $\tau_{py}$, which were used as the input to the command speed generation module in HACS implementation. 
An inertia measurement unit (IMU) attached to rider’s chest recorded the torso yaw angle, $\zeta_z$, which was used to control the ballbot spin. This setup allowed the rider to use the 3-degree-of-freedom (DOF) motion of the torso to control the 3-DOF movement of PURE.

\begin{figure}[thpb]
  \centering
  \includegraphics[scale=0.45]{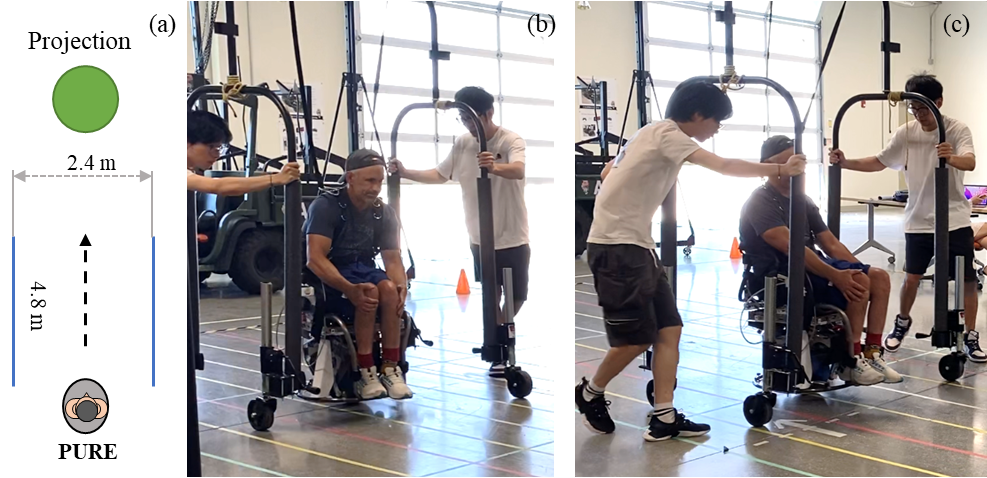}
  \caption{(a) Course layout for the braking test with a projected indicator. (b)–(c) An experienced mWCU riding PURE during the braking test.}
  \label{exp_setup}
\end{figure}

\subsection{Test with Experienced Users}
The first study involved two experienced PURE users from the investigation team (AWB, CX), who each had more than five hours using the HICS and HACS. One participant is a manual wheelchair user (male, 47 yrs, height 1.74 m, weight 63 kg) and the other one is an able-bodied individual (male, 30 yrs, height 1.8 m, weight 60 kg). The testing protocol of this study was approved by the institutional review board (IRB) at the University of Illinois at Urbana-Champaign (UIUC). 

\subsubsection{Experiment Protocol}

A torso function assessment was first conducted to evaluate the torso movement capability of each participant, recording the torso ROM (difference between maximum and minimum torso lean angle) and maximum applied torque using TES when the PURE was parked. A test course was established with 2.4 m width and 4.8 m length defined by tape on the floor (Fig. \ref{exp_setup}a). An overhead projector was used to cast a circular indicator (1.2 m diameter) on the floor at the end of the course. The indicator was green and turned red once the speed of PURE exceeded 1.4 m/s. Participants were instructed to accelerate at a self-selected rate when the indicator was green and to brake with a minimal distance when the indicator turned red. Three trials for each control scheme were collected. To ensure participant safety, they were attached to a mobile gantry system pushed by two researchers (Fig. \ref{exp_setup}b). Participants received ample training sessions to practice hands-free control of PURE and to adjust the sensitivity parameter for each hands-free control scheme. A post-experiment questionnaire was also conducted to assess the preferred hands-free control scheme.

\subsubsection{Data Collection and Processing} 

Torso kinematics ($\zeta_y, \dot\zeta_y$), PURE states and inputs ($\mathbf{s_y}$ and $\tau_y$), and the physical human-robot interactions ($\mathbf{u_{py}}$) were collected using the TES and drivetrain sensors during the torso function assessment and braking experiments ($y$ is associated with the sagittal plane). The assessment results were first processed to obtain the torso ROM ($\zeta_{y_{ROM}}$) and maximum pHRI torque ($\tau_{py_{max}}$) in the sagittal plane. In the braking experiment, the braking effort, resultant torso ROM, maximum applied interaction torque, braking distance, and braking time were calculated for each trial and averaged for each control scheme. For the braking effort estimation (\ref{eqn_brake}), the weighted matrices $\mathbf{Q_y}$ and $\mathbf{Ry}$ were customized for the individual participant based on the torso function assessment results using (\ref{eqn_weight}).

\subsubsection{Results and Discussion}

\begin{figure}[thpb]
  \centering
  \includegraphics[scale=0.5]{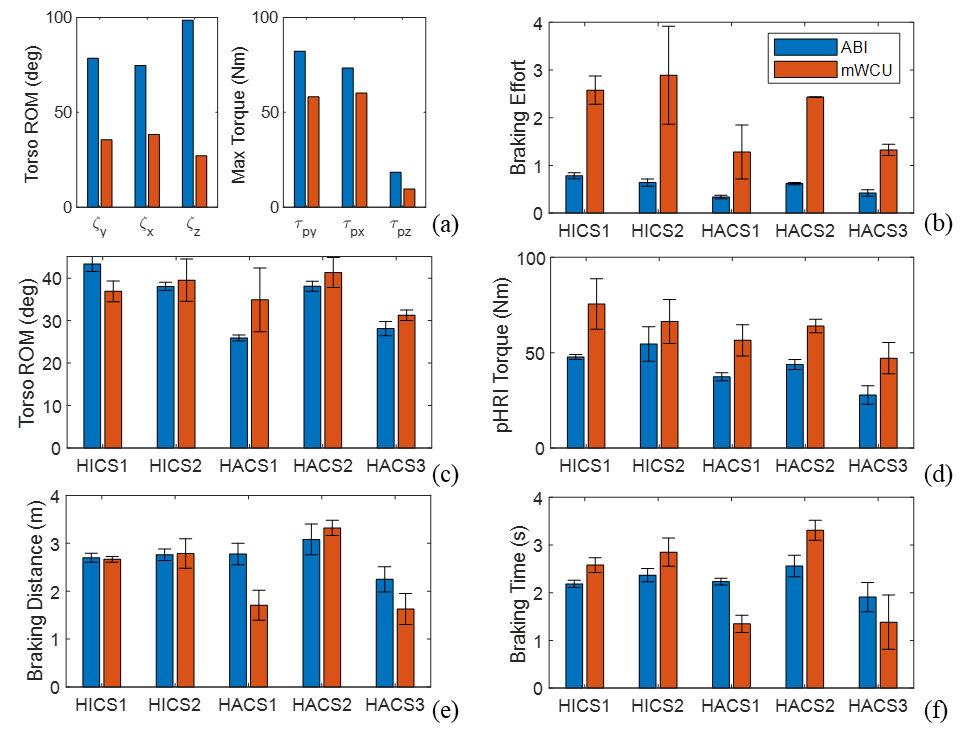}
  \caption{(a) Measured torso range of motion and maximum applied torque to TES by two pilot test participants. (b)-(f) Braking effort, torso ROM, pHRI torque, braking distance, and braking time using five different hands-free control interfaces.}
  \label{braking_exp}
\end{figure}

Despite differences in their torso ROM and maximum pHRI torque (Fig. \ref{braking_exp}a), both participants successfully completed the braking task using all five hands-free control schemes (Fig. \ref{braking_exp}b-f). Using HACS-1 resulted in the lowest braking effort, torso ROM, pHRI torque for both participants, followed by HACS-3. Among HACSs, HACS-2 with pure integration of interaction torque resulted in a higher braking effort for both participants. In general, using HACS-1 and HACS-3 resulted in the most effective braking performance.
In the post-study questionnaire, both participants expressed a preference for HACS-3 in hands-free control of PURE. They found it provided a similar braking performance and provided a smoother riding experience with lower chattering of the system compared with HACS-1. Consequently, the HACS-3 was selected for the second physical robot study to benchmark the braking performance and demonstrate the feasibility of indoor navigation.

\subsection{Test with Novice Users}

Six manual wheelchair users (3M, 3F) and six abled-bodied individuals (3M, 3F) with little prior experience in PURE riding were recruited to benchmark their braking performance with HACS-3 and demonstrate indoor navigation capabilities in two tests (braking, navigation). This study was approved by the IRB at UIUC.

\begin{figure}[thpb]
  \centering
  \includegraphics[scale=0.5]{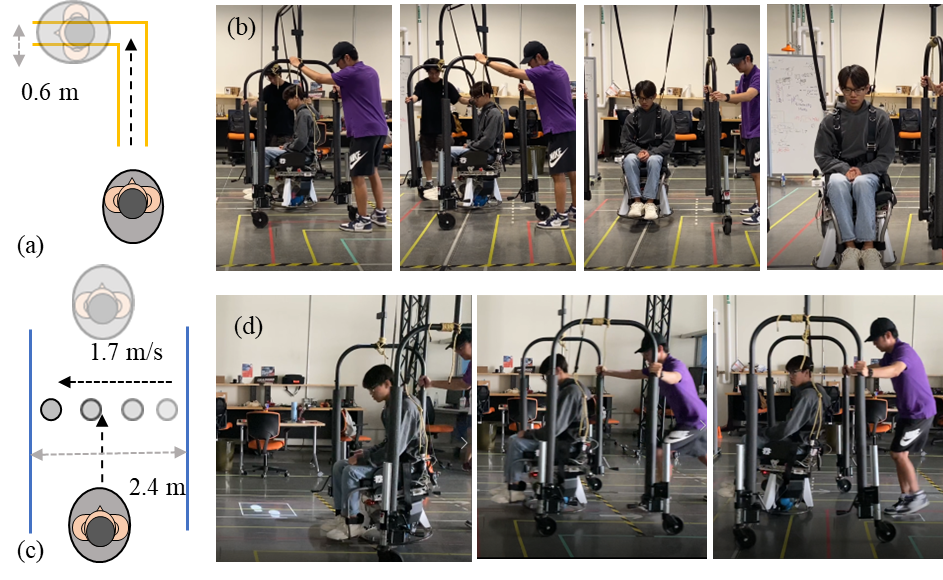}
  \caption{(a)-(b) Left turn course with extreme-narrow width. (c)-(d) Obstacle course with a projected image moving at 1.7 m/s back and forth.}
  \label{nav_test}
\end{figure}

\begin{table}[h]
\caption{Results for Braking Test in Physical Robot Study }
\label{table_results_exp2}
\begin{center}
\begin{tabular}{c c c}
Metrics & ABIs & mWCUs \\
\hline
Braking Effort (J)         & 0.16 $\pm$ 0.03 & 0.27 $\pm$ 0.05\\
Torso ROM  ($^\circ$) & 33.9 $\pm$ 6.0  & 23.1 $\pm$ 6.1\\
Max pHRI Torque (Nm)       & 69.3 $\pm$ 4.9  & 55.2 $\pm$ 5.1\\
Braking Distance  (m)      & 2.39 $\pm$ 0.41 & 2.65 $\pm$ 0.37\\
Braking Time (sec)         & 2.30 $\pm$ 0.55 & 2.68 $\pm$ 0.33\\
\end{tabular}
\end{center}
\end{table}

In the braking test, the sensitivity parameters in HACS-3 were tuned for each participant. A training session ($\sim$30 min) is provided to each rider to practice torso lean control of translational and spinning motions. The same test protocol as in the pilot study was used to obtain the braking performance. In general, these participants were able to brake within an average distance of $2.51$ m and an average time of $2.54$ s from a translational speed of $1.4$ m/s. Higher braking effort ($0.27$), longer braking distance ($2.65$ m), longer braking time ($2.68$ s), smaller torso ROM ($23.1^\circ$), and smaller interaction torque (55.16 Nm) were observed in mWCU compared with ABI participants (TABLE \ref{table_results_exp2}).

Subsequently, these novice users also participated in an indoor navigation test. Testing courses were created using masking tape to replicate various scenarios, including translating forward, turning left and right with various hallway widths, translating sideways, and avoiding static and moving obstacles (Fig. \ref{nav_test}a). Details of the experiment protocol can be found in \cite{song2024driving}. All participants successfully completed all navigation tasks, despite small numbers of retries in some more challenging tasks such as hallway with extreme-narrow width (0.6 m) and passing through moving obstacles (projected on the floor at 1.7 m/s) (Fig. \ref{nav_test}b). More details of the experiment results can be found in \cite{song2024driving}. In general, using HACS-3 with tunable parameters tailored for individual riders allowed them to perform various indoor navigation tasks that have high precision and agility requirements.

\section{DISCUSSION}

In the simulation study, the use of speed command in HACS helped to bring collaborative response into the system dynamics with tunable parameters. In this mode, $\nu_P = 0$ resulted in a similar system response as the baseline LQR-PI controller. As $\nu_P$ increased, the steady-state chassis tilt angle decreased, whereas the braking effort, torso lean angle, and torso torque continued to decrease (Fig. \ref{sim_result}a). With increased $\nu_P$, the ballbot drivetrain began to assist the rider to help maintain the steady-state speed and to assist the braking by tilting the chassis backward once it sensed the applied interaction torque. In addition, HACS provided a more intuitive method to adjust the system behavior based on the torso motion capability of the individual rider via the tunable parameters.

Results from the pilot expert braking study found improved effectiveness of using pHRI information to enable intuitive hands-free control of the translational motions of PURE. The mWCU participant had limited torso ROM compared with the ABI participant (Fig. \ref{braking_exp}a). However, the mWCU participant was still capable of fully braking and achieved a shorter braking distance and time compared with the ABI participant. The relatively worse braking performance of the ABI participant can be reflected by the larger braking effort, torso lean angle, and HRI torque. Provided with enough braking effort, the ABI participant should be able to achieve similar braking distance and time. Nevertheless, this experiment validates the effectiveness of using HACS-1 and HACS-3 to perform agile braking when riding PURE. 

In the novice benchmark study, these less experienced participants achieve similar braking performance ($T_b = 2.51$ s, $ L_b = 2.54$ m) compared with experienced riders ($T_b = 1.8$ s, $L_b = 2.1$ m). It is noteworthy that riders with various sizes and weights adapted well to the fixed-gain LQR-PI balancing controller and two adjustable parameters associated with the speed command generation of HACS-3. The optimal control gains for the LQR-PI controller were developed based on a 60 kg, 1.8 m able-bodied individual. In the current study, weights and heights varied from 50 to 79 kg and 1.60 to 1.78 m. With a weight difference of 36\% and distinct torso motion capabilities, all participants were able to complete the braking task. These results also highlight the intuitiveness of the ballbot riding, as minimal training was required to master the control of the omnidirectional self-balancing mobility device using pure torso motions. In additionally, these less experienced participants were also capable of finishing all indoor navigation navigation tasks. They exhibited high accuracy in position control and timing of ballbot dynamics in courses that invovled extreme-narrow hallway and fast-moving obstacles, showcasing the intuitiveness of torso-lean control of PURE using the proposed hands-free control scheme.


\section{CONCLUSION}

In this paper, we present the development and evaluation of hands-free control schemes for our ballbot mobility device, PURE, using torso motions. A torso-motion sensing system was utilized to measure physical human-robot interactions with the drivetrain and torso kinematics. We explored two groups of hands-free control schemes with and without the use of pHRI information, i.e., hands-free impedance and admittance control schemes (HICS and HACS). Five control schemes across these two groups were tested in a simulation experiment and a pilot physical test with two experienced PURE users. Results from these experiments suggested reduced braking effort with the use of hands-free admittance control schemes (HACS). A follow-up novice benchmark test was then conducted with 12 inexperienced participants using HACS3 to benchmark their braking performance and evaluate feasibility of indoor navigation with challenging scenarios. Their braking performance demonstrated similar braking distance and time compared with both experienced PURE users, and all participants successfully finished all navigation courses. These experiments validated the feasibility of utilizing a ballbot as a drivetrain to provide assistive mobility to riders. They also highlighted the promising result of exploiting physical human-robot interaction for the hands-free control of a riding ballbot with tunable parameters to provide a personal unique rolling experience.

\addtolength{\textheight}{-0cm}   




\section*{ACKNOWLEDGMENT}

The authors thank Doctor Jeannette Elliot, Professor Deana McDonagh, Doctor Patricia Malik, graduate student Nadja Marin, and undergraduate students Chentai Yuan, Yixiang Guo for their help and support with concept development and human subject testing.

\bibliographystyle{IEEEtran}
\bibliography{IEEEabrv, reference}

\begin{thebibliography}{10}
\providecommand{\url}[1]{#1}
\csname url@samestyle\endcsname
\providecommand{\newblock}{\relax}
\providecommand{\bibinfo}[2]{#2}
\providecommand{\BIBentrySTDinterwordspacing}{\spaceskip=0pt\relax}
\providecommand{\BIBentryALTinterwordstretchfactor}{4}
\providecommand{\BIBentryALTinterwordspacing}{\spaceskip=\fontdimen2\font plus
\BIBentryALTinterwordstretchfactor\fontdimen3\font minus \fontdimen4\font\relax}
\providecommand{\BIBforeignlanguage}[2]{{%
\expandafter\ifx\csname l@#1\endcsname\relax
\typeout{** WARNING: IEEEtran.bst: No hyphenation pattern has been}%
\typeout{** loaded for the language `#1'. Using the pattern for}%
\typeout{** the default language instead.}%
\else
\language=\csname l@#1\endcsname
\fi
#2}}
\providecommand{\BIBdecl}{\relax}
\BIBdecl

\bibitem{nichols1979wheelchair}
P.~Nichols, P.~Norman, and J.~Ennis, ``Wheelchair user's shoulder? shoulder pain in patients with spinal cord lesions.'' \emph{Scandinavian journal of rehabilitation medicine}, vol.~11, no.~1, pp. 29--32, 1979.

\bibitem{bayley1987wheelchair}
J.~C. Bayley, T.~Cochran, and C.~Sledge, ``The weight-bearing shoulder. the impingement syndrome in paraplegics.'' \emph{The Journal of bone and joint surgery. American volume}, vol.~69, no.~5, pp. 676--678, 1987.

\bibitem{gellman1988wheelchair}
H.~Gellman, I.~Sie, and R.~L. Waters, ``Late complications of the weight-bearing upper extremity in the paraplegic patient.'' \emph{Clinical Orthopaedics and related research}, no. 233, pp. 132--135, 1988.

\bibitem{omeo}
\BIBentryALTinterwordspacing
``Hans-free, self-balancing wheelchair.'' [Online]. Available: \url{https://myomeo.com/}
\BIBentrySTDinterwordspacing

\bibitem{xiao2023design}
C.~Xiao, M.~Mansouri, D.~Lam, J.~Ramos, and E.~T. Hsiao-Wecksler, ``Design and control of a ballbot drivetrain with high agility, minimal footprint, and high payload,'' in \emph{2023 IEEE/RSJ International Conference on Intelligent Robots and Systems (IROS)}.\hskip 1em plus 0.5em minus 0.4em\relax IEEE, 2023, pp. 376--383.

\bibitem{alami2006safe}
R.~Alami, A.~Albu-Sch{\"a}ffer, A.~Bicchi, R.~Bischoff, R.~Chatila, A.~De~Luca, A.~De~Santis, G.~Giralt, J.~Guiochet, G.~Hirzinger \emph{et~al.}, ``Safe and dependable physical human-robot interaction in anthropic domains: State of the art and challenges,'' in \emph{2006 IEEE/RSJ International Conference on Intelligent Robots and Systems}.\hskip 1em plus 0.5em minus 0.4em\relax IEEE, 2006, pp. 1--16.

\bibitem{de2008atlas}
A.~De~Santis, B.~Siciliano, A.~De~Luca, and A.~Bicchi, ``An atlas of physical human--robot interaction,'' \emph{Mechanism and Machine Theory}, vol.~43, no.~3, pp. 253--270, 2008.

\bibitem{hogan1984impedance}
N.~Hogan, ``Impedance control: An approach to manipulation,'' in \emph{1984 American control conference}.\hskip 1em plus 0.5em minus 0.4em\relax IEEE, 1984, pp. 304--313.

\bibitem{ikeura1995variable}
R.~Ikeura and H.~Inooka, ``Variable impedance control of a robot for cooperation with a human,'' in \emph{Proceedings of 1995 IEEE International Conference on Robotics and Automation}, vol.~3.\hskip 1em plus 0.5em minus 0.4em\relax IEEE, 1995, pp. 3097--3102.

\bibitem{agravante2014collaborative}
D.~J. Agravante, A.~Cherubini, A.~Bussy, P.~Gergondet, and A.~Kheddar, ``Collaborative human-humanoid carrying using vision and haptic sensing,'' in \emph{2014 IEEE international conference on robotics and automation (ICRA)}.\hskip 1em plus 0.5em minus 0.4em\relax IEEE, 2014, pp. 607--612.

\bibitem{yang2017interface}
C.~Yang, C.~Zeng, P.~Liang, Z.~Li, R.~Li, and C.-Y. Su, ``Interface design of a physical human--robot interaction system for human impedance adaptive skill transfer,'' \emph{IEEE Transactions on Automation Science and Engineering}, vol.~15, no.~1, pp. 329--340, 2017.

\bibitem{yu2015human}
H.~Yu, S.~Huang, G.~Chen, Y.~Pan, and Z.~Guo, ``Human--robot interaction control of rehabilitation robots with series elastic actuators,'' \emph{IEEE Transactions on Robotics}, vol.~31, no.~5, pp. 1089--1100, 2015.

\bibitem{love2004force}
L.~J. Love and W.~J. Book, ``Force reflecting teleoperation with adaptive impedance control,'' \emph{IEEE Transactions on Systems, Man, and Cybernetics, Part B (Cybernetics)}, vol.~34, no.~1, pp. 159--165, 2004.

\bibitem{duchaine2007general}
V.~Duchaine and C.~M. Gosselin, ``General model of human-robot cooperation using a novel velocity based variable impedance control,'' in \emph{Second Joint EuroHaptics Conference and Symposium on Haptic Interfaces for Virtual Environment and Teleoperator Systems (WHC'07)}.\hskip 1em plus 0.5em minus 0.4em\relax IEEE, 2007, pp. 446--451.

\bibitem{ficuciello2014cartesian}
F.~Ficuciello, A.~Romano, L.~Villani, and B.~Siciliano, ``Cartesian impedance control of redundant manipulators for human-robot co-manipulation,'' in \emph{2014 IEEE/RSJ International Conference on Intelligent Robots and Systems}.\hskip 1em plus 0.5em minus 0.4em\relax IEEE, 2014, pp. 2120--2125.

\bibitem{ficuciello2015variable}
F.~Ficuciello, L.~Villani, and B.~Siciliano, ``Variable impedance control of redundant manipulators for intuitive human--robot physical interaction,'' \emph{IEEE Transactions on Robotics}, vol.~31, no.~4, pp. 850--863, 2015.

\bibitem{gui2017toward}
K.~Gui, H.~Liu, and D.~Zhang, ``Toward multimodal human--robot interaction to enhance active participation of users in gait rehabilitation,'' \emph{IEEE Transactions on Neural Systems and Rehabilitation Engineering}, vol.~25, no.~11, pp. 2054--2066, 2017.

\bibitem{li2018physical}
Z.~Li, B.~Huang, Z.~Ye, M.~Deng, and C.~Yang, ``Physical human--robot interaction of a robotic exoskeleton by admittance control,'' \emph{IEEE Transactions on Industrial Electronics}, vol.~65, no.~12, pp. 9614--9624, 2018.

\bibitem{kazerooni1993human}
H.~Kazerooni and J.~Guo, ``Human extenders,'' 1993.

\bibitem{colgate2003intelligent}
J.~E. Colgate, M.~Peshkin, and S.~H. Klostermeyer, ``Intelligent assist devices in industrial applications: a review,'' in \emph{Proceedings 2003 IEEE/RSJ International Conference on Intelligent Robots and Systems (IROS 2003)(Cat. No. 03CH37453)}, vol.~3.\hskip 1em plus 0.5em minus 0.4em\relax IEEE, 2003, pp. 2516--2521.

\bibitem{li2012passivity}
Z.~Li, N.~G. Tsagarakis, and D.~G. Caldwell, ``A passivity based admittance control for stabilizing the compliant humanoid coman,'' in \emph{2012 12th IEEE-RAS International Conference on Humanoid Robots (Humanoids 2012)}.\hskip 1em plus 0.5em minus 0.4em\relax IEEE, 2012, pp. 43--49.

\bibitem{augugliaro2013admittance}
F.~Augugliaro and R.~D'Andrea, ``Admittance control for physical human-quadrocopter interaction,'' in \emph{2013 European Control Conference (ECC)}.\hskip 1em plus 0.5em minus 0.4em\relax IEEE, 2013, pp. 1805--1810.

\bibitem{wang2015stability}
H.~Wang, F.~Patota, G.~Buondonno, M.~Haendl, A.~De~Luca, and K.~Kosuge, ``Stability and variable admittance control in the physical interaction with a mobile robot,'' \emph{International Journal of Advanced Robotic Systems}, vol.~12, no.~12, p. 173, 2015.

\bibitem{keemink2018admittance}
A.~Q. Keemink, H.~van~der Kooij, and A.~H. Stienen, ``Admittance control for physical human--robot interaction,'' \emph{The International Journal of Robotics Research}, vol.~37, no.~11, pp. 1421--1444, 2018.

\bibitem{xiao2022personal}
C.~Xiao, ``Personal unique rolling experience: Design, modeling, and control of a riding ballbot,'' Ph.D. dissertation, University of Illinois at Urbana-Champaign, 2022.

\bibitem{spong2004robot}
M.~W. Spong, S.~Hutchinson, and M.~Vidyasagar, ``Robot dynamics and control,'' 2004.

\bibitem{mcneill2002energetics}
R.~McNeill~Alexander, ``Energetics and optimization of human walking and running: the 2000 raymond pearl memorial lecture,'' \emph{American journal of human biology}, vol.~14, no.~5, pp. 641--648, 2002.

\bibitem{kelly2017introduction}
M.~Kelly, ``An introduction to trajectory optimization: How to do your own direct collocation,'' \emph{SIAM Review}, vol.~59, no.~4, pp. 849--904, 2017.

\bibitem{song2024driving}
S.~Y. Song, N.~Marin, C.~Xiao, M.~Mansouri, J.~Ramos, Y.~Chen, A.~W. Bleakney, D.~C. Mcdonagh, W.~R. Norris, J.~R. Elliott \emph{et~al.}, ``Driving a ballbot wheelchair with hands-free torso control,'' in \emph{Proceedings of the 2024 ACM/IEEE International Conference on Human-Robot Interaction}, 2024, pp. 678--686.

\end{thebibliography}

\end{document}